\documentclass[11pt,a4paper]{article}
\usepackage[hyperref]{acl2020}
\usepackage{times}
\usepackage{booktabs}
\usepackage{graphicx}
\usepackage{latexsym}

\usepackage{microtype}

\aclfinalcopy 

\newcounter{fpCounter}
\newif\iffpvar
\fpvartrue
\iffpvar
\newcommand{\fabio}[1]{{\small \color{blue} \refstepcounter{fpCounter}\textsf{[FP]$_{\arabic{fpCounter}}$:{#1}}}}
\else
\newcommand{\fabio}[1]{}
\fi

\newcounter{giCounter}
\newif\ifgivar
\givartrue
\ifgivar
\newcommand{\gautier}[1]{{\small \color{teal} \refstepcounter{giCounter}\textsf{[GI]$_{\arabic{giCounter}}$:{#1}}}}
\else
\newcommand{\gautier}[1]{}
\fi

\title{A Memory Efficient Baseline for Open Domain Question Answering}

\author{
  Gautier Izacard$^{1,2,3}$ \hspace{1em}
  Fabio Petroni$^1$ \hspace{1em}
  Lucas Hosseini$^1$ \AND  
  Nicola De Cao$^4$ \hspace{1em}
  Sebastian Riedel$^{1,5}$ \hspace{1em} 
  Edouard Grave$^1$ \vspace{0.1cm} \\
  $^1$ Facebook AI Research \hspace{.5em}
  $^2$ ENS, PSL University \hspace{.5em} 
  $^3$ Inria \hspace{.5em} \\
  $^4$ University of Amsterdam \hspace{.5em}
  $^5$ University College London \\
  \texttt{gizacard|fabiopretroni|hoss|ndecao|sriedel|egrave@fb.com} \\
}

\date{}

\begin{document}
\maketitle
\begin{abstract}
Recently, retrieval systems based on dense representations have led to important improvements in open-domain question answering, and related tasks.
While very effective, this approach is also memory intensive, as the dense vectors for the whole knowledge source need to be kept in memory.
In this paper, we study how the memory footprint of dense retriever-reader systems can be reduced.
We consider three strategies to reduce the index size: dimension reduction, vector quantization and passage filtering.
We evaluate our approach on two question answering benchmarks: TriviaQA and NaturalQuestions, showing that it is possible to get competitive systems using less than 6Gb of memory.
\end{abstract}

\section{Introduction}


The goal of open-domain question answering is to answer factoid questions about general topics~\citep{voorhees1999trec}.
Since the evidence to answer the question is not provided to the system, a standard approach is to use an external source of knowlege such as Wikipedia.
These systems, known as \emph{retriever-reader}, start by retrieving relevant documents, or passages, from the knowledge source, and then process them with a reader model to produce the answer~\citep{chen2017reading}.

The downstream performance of these pipeline systems greatly depends on the quality of the retrieval module.
Traditionally, queries and documents were represented as sparse vectors based on hand crafted term weighting such as BM25~\citep{robertson1995okapi}.
More recently, dense representations obtained with neural networks have proven very effective for question answering, and related tasks.
A limitation of these methods is the size of the index, which can be tens of gigabytes, as all the vector representations of the knowledge source have to be kept in memory.
This lead to the following question: \emph{what is the more efficient way to store information for question answering?}

In this paper, we explore how non-parametric approaches to store information can be used for memory efficient question answering systems.
More precisely, starting from a state-of-the-art pipeline using a dense retriever, we study how the size of the index can be reduced.
We consider three main strategies to do so: dimension reduction of the dense representations, vector quantization and document filtering.
We study the impact of these three techniques on the downstream performance, using two open-domain question answering benchmarks: TriviaQA and NaturalQuestions.
We show that, with minimal loss of performance, it is possible to obtain QA systems using less than 6Gb of memory.



\section{Related Work}

\paragraph{Retriever-Reader Approach.}
Given a question, retriever-reader systems first retrieve relevant documents from an external knowledge source~\citep{chen2017reading}.
These documents, along with the question, are then processed by a second model called the reader to produce the answer.
In case where no gold spans are available, \citet{clark2017simple} proposed a loss function based on global normalization over all the mentions of the answer.
\citet{wang2019multi} applied this approach to BERT pre-trained models, while \citet{min2019discrete} introduced an alternative method based on expectation-maximization.
\citet{roberts2020much} used large pre-trained generative models, generating the answer without retrieving external documents, a setting referred to as \emph{closed book}.
Several works showed that generative QA models could be improved by retrieval, using the obtained documents as input of the sequence-to-sequence model~\citep{min2020ambigqa,lewis2020retrieval,izacard2020leveraging}.

\paragraph{Dense Embedding Retrieval.}
Traditional information retrieval is based on sparse vector representations, with hand crafted term weighting~\citep{jones1972statistical,robertson1995okapi}.
Given a query, a similarity score such as the dot product is used to retrieve the most relevant documents.
Recently, it has been shown that dense representations, obtained with neural networks, were a good alternative to sparse representations.
Different training schemes of the retriever have been considered, based on supervised learning~\citep{karpukhin2020dense}, latent modeling~\citep{lee2019latent,guu2020realm} or distillation~\citep{izacard2020distilling}.
Finally, \citet{luan2020sparse} studied the impact of design choices, such as document length or vector dimension, on the retrieval performance.
We differ from this work by evaluating on the downstream QA task, and exploring other techniques like quantization and document filtering.

\paragraph{Product Quantization.}
The process of quantization~\citep{gray1998quantization} consists in mapping values from a continuous or large space (eg. \texttt{float32}) to a smaller discrete set (eg. \texttt{int8}).
This process is the basis of most lossy data compression techniques.
In vector, or block, quantization, all the elements of a vector are quantized simultaneously instead of independently.
One of the simplest vector quantization method is to use $k$-means, and map each vector to the closest centroid.
A better approach in high-dimension, known as product quantization~\citep{jegou2010product}, is to subdivide each vector of dimension $d$ into $n$ sub-vectors of dimension $d/n$ and to quantize these independently, using $k$-means.

\section{System description}
We briefly describe our topline QA system, following the retriever-reader approach. 


For the reader we use a Fusion-in-Decoder model~\citep{izacard2020leveraging}. 
This type of model leverages the sequence-to-sequence architecture to efficiently combine multiple passages in order to generate an answer.
In the Fusion-in-Decoder model, the encoder of the sequence-to-sequence model is first applied independently on each passage concatenated with the question.
Then, the representations are concatenated and the decoder is applied on the resulting representations to predict an answer.
We initialize the Fusion-in-Decoder model with T5~\citep{raffel2019exploring}. 

For the retriever, we use an approach based on dense embeddings~\citep{karpukhin2020dense}.
First, each passage $p$ of the knowledge source is mapped to a $d$-dimensional vector using an embedding function $E$.
Then, given an input question $q$, it is represented with the same embedding function.
A similarity score between passage $p$ and question $q$ is obtained by the dot product between the two representations $S(q, p) = E(q)^T E(p)$.
Passages with the highest similarity scores are then retrieved and processed by the reader.
This operation is done efficiently by using a maximum inner product search library such as Faiss~\cite{johnson2019billion} after pre-indexing all Wikipedia passages.

The retriever is trained with knowledge distillation, where the synthetic labels are obtained by aggregating the attention scores of the reader model~\citep{izacard2020distilling}.
We start by training a reader with the passages obtained with DPR~\citep{karpukhin2020dense}.
Then, a new retriever is obtained by distilling the aggregated cross-attention scores of the reader to the retriever.
Finally, new passages are obtained with this retriever, and used to train a new reader model.


\section{Compressing QA systems}
In this section, we discuss the size of our topline system, and techniques to reduce it.

\subsection{Initial system size}
\paragraph{Models.}
Our system uses two neural models, the retriever and the reader.
The retriever is based on the BERT base architecture, containing 110M parameters and thus weighting 0.22Gb in \texttt{float32}.
The reader is based on T5 base and large, containing 220M and 770M parameters respectively, thus weighting 0.88Gb and 3.1Gb in \texttt{float32}.

\paragraph{Knowledge source.}
Our system follow standard open-domain QA practice and uses Wikipedia as knowledge source.
We use the English Wikipedia\footnote{Under https://creativecommons.org/licenses/by-sa/3.0/} dump from Dec. 20, 2018, including lists, which weighs approximately 21Gb uncompressed.
This can be reduced to 3.8Gb by using the compression softward \texttt{xz}, or even 3.1Gb with \texttt{lrzip}.



\begin{figure*}[t]
\begin{center} 
\includegraphics{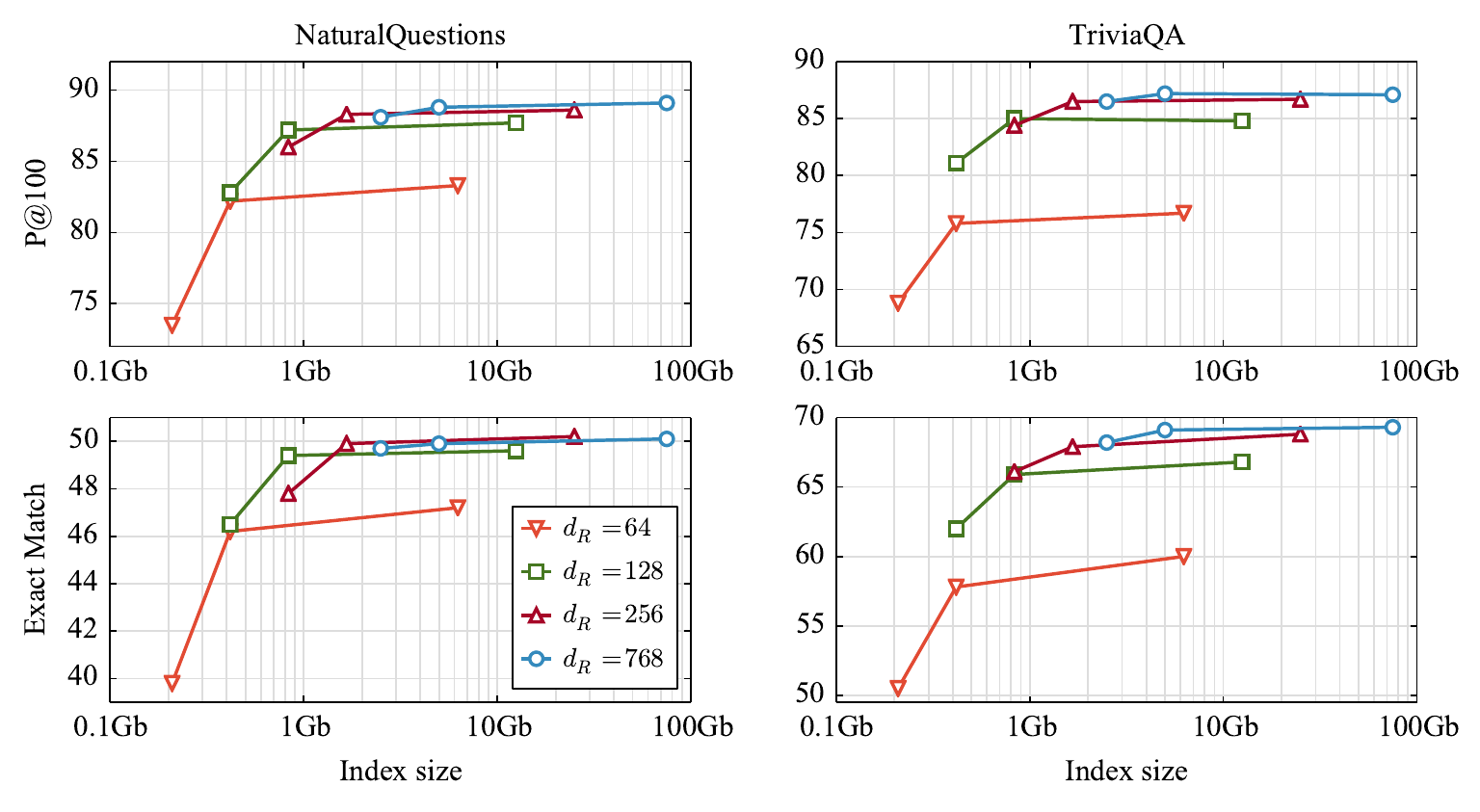}
\caption{Perfomance as a function of index size, for different dimension and quantization.
  Each curve corresponds to a fixed dimension, the left point using $1$ bit per dimension, the middle point $2$ bit and the right point $32$ bit.
}
\label{fig:tradeoff}
\end{center}
\end{figure*}

\paragraph{Dense index.}
Both dense and sparse representations lead to a large index, using a significant memory footprint.
Following \citet{karpukhin2020dense}, we split Wikipedia into non-overlapping chunks of 100 words.
Then, each chunk is represented by a vector of dimension 768 obtained with the retriever model.
Assuming that vectors are stored in \texttt{float32}, this leads to an index size of 75Gb.


\subsection{Reducing the index size}
In the following, we discuss strategies to reduce the size of QA systems.
As noted previously, most of the memory footprint is allocated to the index, and we will thus focus on reducing its size.

\paragraph{Dimension reduction.}
One of the simplest way to reduce the index size is to reduce the dimension $d$ of the dense representations of passages.
We consider to reduce the embedding dimension at train time, by adding a linear layer that maps the output of the network to a $d_R$-dimensional vector, followed by a LayerNorm.
This follows the same solution adopted in~\cite{luan2020sparse}, where the trade-off between the dimension reduction and the retrieval performance is studied. 
We empirically verified that adding the LayerNorm led to stronger performance than without it.
An alternative strategy would be to use principal component analysis on the set of already trained embeddings.

\paragraph{Product quantization.}
A second way to reduce the index size is to use vector quantization, which is complementary to dimension reduction.
Standard vectors are stored in memory by using $4$~bytes per dimension when using \texttt{float32}.
Without loss of performance, it is possible to use \texttt{float16}, leading to a compression factor of two.
To compress representations more aggressively, we propose to use product quantization.
In that case, embeddings are divided in $n_v$ sub-vectors of dimension $d/n_v$, each being stored using $n_b$ bits.
If we use sub-vectors of dimension two and use one byte to store them, then, each dimension is using $0.5$~bytes.
This leads to a compression factor of eight compared to the original index.


\paragraph{Passage filtering.}
A last method to reduce the index size is to remove documents that are unlikely to be useful for question answering.
To do so, we train a linear classifier, where each Wikipedia article is represented by its title and list of categories.
As positive examples, we use articles retrieved with DPR on the training data.
To obtain negative examples, we use self-training: we start by randomly sampling Wikipedia articles as negatives, and train a first classifier.
Then, the articles classified as negative with the highest confidence are used as negatives for the next iteration of self-training.
We perform a few iterations of this scheme, and apply the final classifier to filter the Wikipedia dump.


\section{Experiments}
We conduct experiments on two datasets: NaturalQuestions~\cite{kwiatkowski2019NQ} and TriviaQA~\cite{JoshiTriviaQA2017} and follow the standard setting for open-domain QA~\cite{karpukhin2020dense}.
The final end-to-end performance of the system is evaluated using the Exact Match score.
We also report the top-k retrieval accuracy (P@k): the percentage of questions for which at least one of the top-k retrieved passages contains the gold answer.

\subsection{Technical details}

\begin{figure}[t]
\begin{center} 
\includegraphics{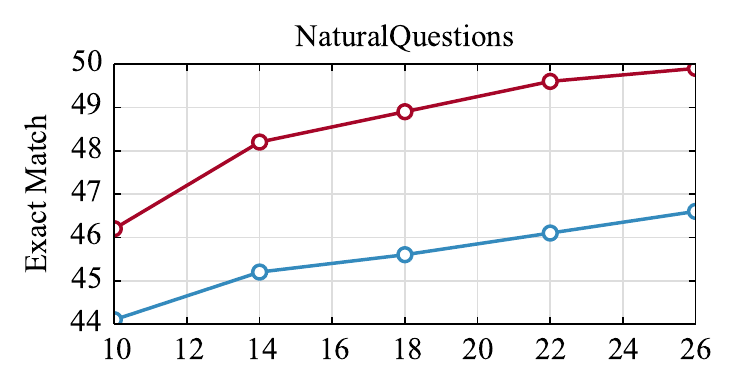}
\caption{Perfomance as a function of the number of passages, in millions (full wikipedia contains 26M).
}
\label{fig:passage}
\end{center}
\end{figure}

\paragraph{Retriever.}
We initialize the retriever with the uncased BERT base model~\citep{devlin2018bert}, and use the same model to embed questions and passages.
The representations of the final layer are averaged instead of using the representation corresponding to the \texttt{[CLS]} token.
Furthemore, similarly to~\citet{khattab2020relevanceguided}, we pad questions and passages to fixed-length sequences without using mask when applying the embedder function.
We use sentence of length 40 tokens for the questions, and 200 tokens for the passages.

\paragraph{Reader.}
The reader model is initialized with a T5-base model unless otherwise specified.
We use 100 retrieved passages per question to train the model, using AdamW~\cite{loshchilov2018decoupled} with a batch size of 64.
We train the model with a peak learning rate of $10^{-4}$.
The learning rate increases linearly during 600 gradient steps, and decreases linearly during 14.4k steps. 
The best model is selected based on the exact match score evaluated on the dev set every 500 gradient steps.

\subsection{Dimension reduction vs. quantization}
We start by investigating if dimension reduction and product quantization are complementary, and what setting works best.
For this, we train retriever-reader systems with different dimensions $d_R$ of the index. 
Then, we use product quantization to compress the pre-computed index.
The number of bits per subvector is fixed at 8, and we vary the number of subvectors.
Results are reported in Fig.~\ref{fig:tradeoff}.
It appears that dimension reduction and quatization are complementary, as pareto optimal systems often involve both.
Combining them allows to obtain an index of reasonable size, 1.6Gb, with little accuracy loss: -0.2EM on NQ and -1.1EM on TriviaQA.

\subsection{Passage filtering}
Filtering passages allows to reduce the size of both the index and the knowledge source.
In Figure~\ref{fig:passage} we progressively discard articles starting from the full Wikipedia containing 26M passages. 
We report results using two retriever-reader pipelines trained in the previous subsection with $d_R=256$ and $d_R=128$.
For the first one we use a quantized index with 64 subvectors per passage, this leads to an index of size 1.67Gb for the complete Wikipedia.
The second uses 16 subvectors per passage and its original size on the whole Wikipedia is 0.42Gb.
As shown here it is possible to discard a significant fraction of Wikipedia articles.

\subsection{Final models}
Finally, we report in Table~\ref{tab:sota} the test performance of two compressed systems, and compare them to our state-of-the-art topline models.
The first system, at 5.1Gb, is made of a reader large, index of dimension 256 with 2 bits per dimension and 18M of passages.
The second system, at 2.1Gb, is made of a reader small, index of dimension 128 with with 2 bits per dimension and 10M passages.
We observe that these systems, while significantly smaller than the topline, obtain competitive performance.

\begin{table}
\centering
\begin{tabular}{llcc}
  \toprule
  Model & size & NQ & TriviaQA \\
  \midrule
  Topline base & 79Gb & 50.4 & 69.8 \\
  Topline large & 81Gb & 54.7 & 73.3 \\
  \midrule
  Compressed base & 2.1Gb & 44.0 & 56.8 \\
  Compressed large & 5.1Gb & 53.6 & 71.3 \\
  \bottomrule
\end{tabular}
\caption[Caption]{Test set accuracy of compressed systems and our topline models.}
\label{tab:sota}
\end{table}

\section*{Discussion}
In this paper, we explored how to compress retriever-reader pipelines based on dense representations.
We studied different strategies, which are complementary, such as dimension reduction, vector quantization and passage filtering.
When combining these different methods, we showed that it is possible to obtain systems smaller than 6Gb, which are competitive with the state-of-the-art.

\bibliographystyle{acl_natbib}
\bibliography{references}

\end{document}